# Geospatial Semantics

Yingjie Hu

GSDA Lab, Department of Geography, University of Tennessee, Knoxville, TN 37996, USA

**Abstract**

Geospatial semantics is a broad field that involves a variety of research areas. The term *semantics* refers to the meaning of things, and is in contrast with the term *syntactics*. Accordingly, studies on geospatial semantics usually focus on understanding the meaning of geographic entities as well as their counterparts in the cognitive and digital world, such as cognitive geographic concepts and digital gazetteers. Geospatial semantics can also facilitate the design of geographic information systems (GIS) by enhancing the interoperability of distributed systems and developing more intelligent interfaces for user interactions. During the past years, a lot of research has been conducted, approaching geospatial semantics from different perspectives, using a variety of methods, and targeting different problems. Meanwhile, the arrival of big geo data, especially the large amount of unstructured text data on the Web, and the fast development of natural language processing methods enable new research directions in geospatial semantics. This chapter, therefore, provides a systematic review on the existing geospatial semantic research. Six major research areas are identified and discussed, including semantic interoperability, digital gazetteers, geographic information retrieval, geospatial Semantic Web, place semantics, and cognitive geographic concepts.

## 1. Introduction

The term *semantics* refers to the *meaning* of expressions in a language, and is in contrast with the term *syntactics*. For example, the two expressions "I love GIS" and "I ❤ GIS" have clearly different syntactics; however, they have close, if not the same, semantics. *Geospatial semantics* adds the adjective *geospatial* in front of *semantics*, and this addition both restricts and extends the initial applicable area of *semantics*. On one hand, geospatial semantics focuses on the expressions that have a connection with geography rather than any general expressions; on the other hand, geospatial semantics enables studies on not only linguistic expressions but also the meaning of geographic places, geospatial data, and the GeoWeb.

_______________________________

While geospatial semantics is a recognized subfield in GIScience (Agarwal, 2005; D. Mark, Egenhofer, Hirtle, & Smith, 2000), it also involves a variety of related research areas. Kuhn (2005) defines geospatial semantics as "understanding GIS contents, and capturing this understanding in formal theories." This definition can be divided into two parts: understanding and formalization. The understanding part triggers the question: who is supposed to understand the GIS content, people or machines? When the answer is "people", research in geospatial semantics involves human cognition of geographic concepts and spatial relations (Egenhofer & Mark, 1995; Golledge, 2002; B. Smith & Mark, 2001); whereas when the answer is "machines", it can involve research on the semantic interoperability of distributed systems, digital gazetteers, and geographic information retrieval (Y. Bishr, 1998; F. T. Fonseca, Egenhofer, Agouris, & Câmara, 2002; Goodchild & Hill, 2008; Harvey, Kuhn, Pundt, Bishr, & Riedemann, 1999; C. B. Jones & Purves, 2008). The second part of the definition proposes to capture this understanding through formal theories. Ontologies, as formal specifications of concepts and relations, have been widely studied and applied in geospatial semantics (Couclelis, 2010; Frank, 2001; Pundt & Bishr, 2002; Visser, Stuckenschmidt, Schuster, & Vögele, 2002), and formal logics, such as first-order logic (Russell, Norvig, Canny, Malik, & Edwards, 2003) and description logics (Hitzler, Krotzsch, & Rudolph, 2009), are often employed to define the concepts and axioms in an ontology. While Kuhn's definition includes these two parts, research in geospatial semantics is not required to have both-- one study can focus on *understanding*, while another one examines *formalization*.

Advances in computer and information technologies, especially the Web, have greatly facilitated geospatial semantic research. With the Semantic Web initially proposed by Berners-Lee, Hendler, and Lassila (2001), Egenhofer (2002) envisioned the Geospatial Semantic Web which is able to understand the semantics of geospatial requests of users and automatically obtain relevant results. The development of Linked Data (Bizer, Heath, & Berners-Lee, 2009) as well as the resulting Linked Open Data cloud (Heath & Bizer, 2011) have fostered geospatial semantic studies on organizing, publishing, retrieving, and reusing geospatial data as structured Linked Data (Janowicz, Scheider, & Adams, 2013). Meanwhile, there is a rapid increase in the volume of unstructured natural language text on the Web, such as social media posts, blogs, and Wikipedia entries. While often subjective, textual data reveal the understanding and perceptions of people towards natural and social environments. Existing studies have demonstrated the use of unstructured text data in extracting place semantics and understanding the spatiotemporal interaction patterns between people and places (Adams & McKenzie, 2013; Ballatore & Adams, 2015; Y. Hu, McKenzie, Janowicz, & Gao, 2015). More novel research topics based on big text data will become possible, given the fast development of natural language processing (NLP) and text mining techniques.

Geospatial semantics is a broad field that adopts a unique research perspective towards geospatial problems. To some extent, geospatial semantics can be compared with geospatial statistics: both can be applied to various problems across domains and both have their own unique set of methods (e.g., ontological modeling and natural language processing for geospatial semantics). In recent years, a lot of research on geospatial semantics has been conducted, and the results are published in journals or presented in conferences, such as the Conference on Spatial Information Theory (COSIT), the International Conference on Geographic Information Science (GIScience), the International Conference on Geospatial Semantics (GeoS), and many others. This chapter systematically reviews and summarizes the existing efforts. The objective is to delineate a road map that provides an overview on six major research areas in geospatial semantics.

## 2. Six Major Research Areas

### 2.1 Semantic Interoperability and Ontologies

Semantic interoperability was driven by the componentization of GIS. While GIS were traditionally used locally, geospatial functions and data were increasingly encapsulated into services and shared on the Web (Kuhn, 2005). As a result, it became necessary to formally define the semantics of the distributed Web services, so that they can automatically interact with each other and be dynamically integrated. Semantic interoperability is also critical for Spatial Data Infrastructures (SDIs) that provide access to a wealth of distributed geospatial data sources and services which can be combined for various queries and tasks (Alameh, 2003; Lemmens, et al., 2006; Michael Lutz, Lucchi, Friis-Christensen, & Ostländer, 2007; Mickey Lutz, Sprado, Klien, Schubert, & Christ, 2009). While semantic interoperability can refer to any match-making process (e.g., matching a Web document to a user's query), this section will focus on enabling the integration among distributed geospatial data and services.

A major approach for enabling semantic interoperability is developing ontologies. While studied in the field of philosophy as the nature of being, ontologies in geospatial semantics are closer to those in computer science and bioinformatics, which serve the function of formalizing the meaning of concepts in a machine-understandable manner (Bittner, Donnelly, & Winter, 2005; Couclelis, 2009; Gruber, 1993; Guarino, 1998; Stevens, Goble, & Bechhofer, 2000). From a data structure perspective, an ontology can be considered a graph with concepts as nodes and relations as edges. Figure 1(a) shows a fragment taken from an example ontology developed in the GeoLink project supported by the U.S. National Science Foundation (Adila Krisnadhi, et al., 2015). The concepts (e.g., *Cruise* and *Vessel*) and the relations (e.g., *isUndertakenBy*) in this example ontology have been labeled using terms in the corresponding domain (oceanography in this case). Ontologies are often embedded into GIS and Web services as an additional component to enable semantic interoperability (Frank, 1997). Examples include the Ontology-Driven GIS (ODGIS) proposed by F. T. Fonseca and Egenhofer (1999), as well as other ontology-based approaches developed by Hakimpour and Timpf (2001), F. T. Fonseca, Egenhofer, Davis, and Borges (2000), and Fallahi, Frank, Mesgari, and Rajabifard (2008). Kuhn (2003) proposed semantic reference systems (SRS), an ontology-based system analogous to the existing spatial and temporal reference systems widely used in GIS to facilitate semantic interoperability.

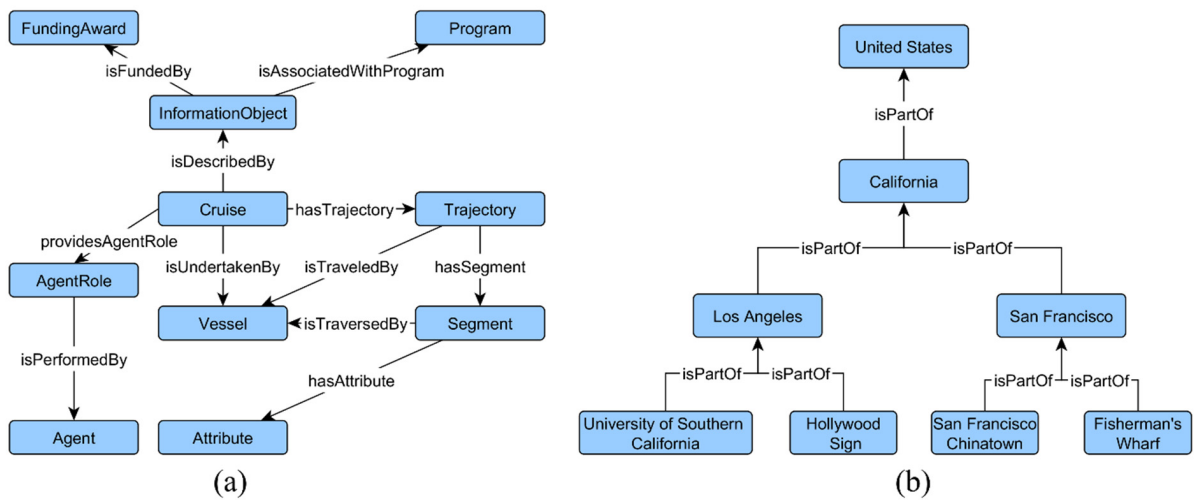

Figure 1. (a) A fragment of an ontology from the NSF GeoLink project (Adila Krisnadhi, et al., 2015); (b) a fragment of a possible gazetteer.

Ontologies must be developed before they can be used in a GIS. Three types of ontologies can be identified from the literature: top-level ontology, domain ontology, and ontology design pattern (ODP). A top-level ontology contains general terms (e.g., *isPartOf, endurant,* and *perdurant*) that can be used across domains, while domain ontologies formalize the concepts for a specific discipline (Ashburner, et al., 2000; Guarino, 1997; Rogers & Rector, 1996). The ontologies used in GIScience are generally considered as domain ontologies, and are often called *geographic ontologies* or *geo-ontologies* (F. Fonseca, Câmara, & Miguel Monteiro, 2006; Tomai & Kavouras, 2004a). Ontology design patterns are developed based on applications. Instead of seeking agreements within or across domains, they capture the common needs shared by multiple applications (Gangemi, 2005; Gangemi & Presutti, 2009). The process of developing ontologies is called ontology engineering. Three types of approaches are often used for ontology engineering, which are top-down, bottom-up, and hybrid approaches. Top-down approaches rely on knowledge engineers and domain experts to define and formalize the ontological concepts and relations (Brodaric, 2004; Gates, Keller, Salayandia, Da Silva, & Salcedo, 2007; Schuurman & Leszczynski, 2006; Shankar, et al., 2007; Y. Wang, Gong, & Wu, 2007). Bottom-up approaches employ data mining methods to extract concepts and relations from structured databases or unstructured natural language text (Baglioni, Masserotti, Renso, & Spinsanti, 2007; Buitelaar, Cimiano, & Magnini, 2005; Maedche & Staab, 2004; Sen, 2007; Shamsfard & Barforoush, 2004). The hybrid approaches integrate the previous two and combine both expert knowledge and results from data mining processes (Buitelaar, Olejnik, & Sintek, 2004; Y. Hu & Janowicz, 2016; Prieto-Díaz, 2003). One challenge in ontology engineering is to define the semantics of the *primitive terms* (the atomic concepts that cannot be further divided) in an explicit and unambiguous manner. To address this challenge, researchers have proposed to ground the primitive terms based on the environment and the observation process (Janowicz, 2012; Mallenby, 2007; Scheider, Janowicz, & Kuhn, 2009; Schuurman, 2005; Third, Bennett, & Mallenby, 2007).

Ontologies can be encoded using formal logics, but can also be implemented as a simple structured vocabulary. A data standard is a simple ontology (Bittner, et al., 2005). So far, many geo-ontologies have been developed. For example, Grenon and Smith (2003) developed SNAP and SPAN which are two general geo-ontologies for modeling *continuants* and *occurrents* respectively. Worboys and Hornsby (2004) proposed the geospatial event model (GEM) which extends the traditional object model with events to capture dynamic geospatial processes. There are also geo-ontologies for ecosystems (Sorokine, Sorokine, Bittner, & Renschler, 2004), hydrology (Feng, Bittner, & Flewelling, 2004), and Earth Science (Raskin & Pan, 2005). In addition, multi-layer ontologies, which distinguish the entities in the physical world and their representations in human cognition, were developed for spatiotemporal databases (Frank, 2001, 2003) and geographic information (Couclelis, 2010). Recent years have witnessed rapid development of ODPs in the geospatial domain, such as the ODPs for semantic sensor networks (Compton, et al., 2012), semantic trajectories (Y. Hu, et al., 2013), barrier dynamics (White & Stewart, 2015), cartographical scale (Carral, et al., 2013), surface water features (Sinha, et al., 2014), oceanographic cruises (AA Krisnadhi, et al., 2015), and space-time prisms (Keßler & Farmer, 2015).

With many ontologies developed by different researchers and communities, it is often necessary to align these ontologies to support data integration, a process known as ontology alignment (Cruz, Sunna, & Chaudhry, 2004; Hess, Iochpe, Ferrara, & Castano, 2007). Based on the alignment direction, we can identify centralized and peer-to-peer alignments (Sunna & Cruz, 2007). The former aligns multiple ontologies to a standard ontology, while the latter establishes links between two peer ontologies. The alignment methods can be classified into: element-level, structure-level, and hybrid methods (Shvaiko & Euzenat, 2005). Element-level methods focus on the individual concepts and relations in an ontology and compare the similarities of the label strings as well as their dictionary definitions (e.g., based on WordNet (Miller, 1995)) (Lin & Sandkuhl, 2008). Structure-level methods examine not only the terms themselves but also their neighboring concepts in the ontology (Sunna & Cruz, 2007). Hybrid approaches combine the element-level and structure-level methods, and examples include the Alignment API (Euzenat, 2004), SAMBO (Lambrix & Tan, 2006), RiMOM (J. Li, Tang, Li, & Luo, 2009), and Falcon-AO (W. Hu & Qu, 2008). There are also studies that align ontologies based on the instances inside the ontology concepts (Brauner, Casanova, & Milidiú, 2007; Navarrete & Blat, 2007). As conflicts can rise during the alignment process, some research has incorporated human experts in the alignment process, such as COMA++ (Aumueller, Do, Massmann, & Rahm, 2005) and AgreementMaker (Cruz, Sunna, Makar, & Bathala, 2007).

While a lot of research has been conducted on ontologies, it is worth noting that developing an ontology is only one approach for realizing semantic interoperability. In fact, some researchers have criticized the use of ontologies to address semantic issues by arguing that ontologies as priori agreements cannot capture the meaning of concepts that change dynamically (Di Donato, 2010; Gärdenfors, 2004). New approaches for semantic interoperability may also be possible and will need further investigation.

## 2.2. Digital Gazetteers

Digital gazetteers are structured dictionaries for named places. The place entries within a digital gazetteer are often organized into a graph, with nodes representing places and edges capturing their relations (e.g., Los Angeles is part of California). In fact, digital gazetteers can be considered as a special type of ontology. Figure 1(b) shows a fragment of a possible gazetteer whose organization structure shares similarity with the ontology fragment in Figure 1(a). The reason that digital gazetteers are discussed as a separate section is their vital importance in GIScience (Goodchild & Hill, 2008). There exist many applications of digital gazetteers, including geocoding, navigation, and geographic information retrieval (Alani, Jones, & Tudhope, 2001; Rice, Aburizaiza, Jacobson, Shore, & Paez, 2012; Schlieder, Vögele, & Visser, 2001). Three core components are usually contained in a digital gazetteer: place names (N), place types (T), and spatial footprints (F) (Hill, 2000). These three components enable three common operations: spatial lookup (N → F), type lookup (N → T), and reverse lookup (F (×T) → N) (Janowicz & Keßler, 2008). As people frequently use place names rather than numeric coordinates to refer to places, digital gazetteers fill the critical gap between informal human discourses and formal geographic representations. From a perspective of geospatial semantics, digital gazetteers help machines understand the geographic meaning (e.g., the spatial footprint) of a textual place name as well as the relations among places. Examples of digital gazetteers include GeoNames, Getty Thesaurus for Geographic Names (TGN), GEOnet Names Server (GNS), Geographic Names Information System (GNIS), and the Alexandria Digital Library Gazetteer (ADL).

One important topic in gazetteer research is enriching existing gazetteers with local or vernacular place entries. Gazetteers are traditionally developed and maintained by naming authorities (e.g., the Board on Geographic Names in the U.S.), and often do not contain the local place names used in everyday conversations (Davies, Holt, Green, Harding, & Diamond, 2009; Hollenstein & Purves, 2010). For example, the entry *San Francisco Chinatown* shown in Figure 1(b) may not be included in a traditional gazetteer. Yet, such vernacular places are important for some GIS applications (e.g., finding the hotels in *San Francisco Chinatown* for tourists). Since these places often do not have clearly defined boundaries, research has been conducted to represent their vague spatial footprints. For example, Burrough and Frank (1996) used a fuzzy-set-based approach to extract the intermediate boundaries of vague places. Montello, Goodchild, Gottsegen, and Fohl (2003) asked human participants to draw the boundary of *downtown Santa Barbara*, and found a common core area agreed upon by the participants. C. B. Jones, Purves, Clough, and Joho (2008) proposed a computational approach which employs a Web search engine to harvest the geographic entities (e.g., hotels) associated with a vague place name, and then used kernel density estimation (KDE) to represent the vague boundary. Geotagged photos (e.g., Flickr photos) provide natural links between textual tags (which often contain vernacular place names) and geographic locations, and have been utilized by many researchers to model vague places, such as Grothe and Schaab (2009), Keßler, Maué, Heuer, and Bartoschek (2009), Intagorn and Lerman (2011), L. Li and Goodchild (2012), and Gao, Li, Li, Janowicz, and Zhang (2014). The methods used for modeling vague boundaries include KDE, characteristic shape (Duckham, Kulik, Worboys, & Galton, 2008), and the 'egg-yolk' representation (Cohn & Gotts, 1996). A recent work from Chen and Shaw (2016) proposed a weighted KDE approach that assigns different weights to Flickr photo locations with different representativeness of the vague place. In addition, there is research that focuses on assigning place types to place instances, rather than generating geometric representations for spatial footprints. For example, Uryupina (2003) developed a bootstrapping approach that can automatically classify places into predefined types (e.g., *city* and *river*), and achieved a high precision of about 85%. It is worth noting that the inclusion of user generated content (e.g., geotagged photos) brings the issues of data quality and credibility. User reputations (M. Bishr & Kuhn, 2007), topological relations (e.g., *islands* should be surrounded by water) (Keßler, Janowicz, & Bishr, 2009), and other methods could be used to address these issues.

Another topic in digital gazetteer research is to align and conflate multiple gazetteers. Digital gazetteers from independent sources may have different geographic coverages, different spatial footprints (e.g., the same place may be represented as a point or as a polygon in different gazetteers), different place types, and different attributes. While these differences can be combined to form a richer data resource, they also present challenges for gazetteer conflation. Based on the conflating targets, we can identify schema-level and instance-level conflations. The schema-level conflation aligns the place types from one gazetteer to that of another gazetteer, and can be considered as a special type of ontology alignment. Naturally, the ontology alignment methods, such as those based on the similarities of labels, definitions, and structures, can be employed to align the place types in different gazetteers (Rodríguez & Egenhofer, 2003). There are also methods that leverage the spatial distribution patterns of place instances belonging to a place type to align place types, such as the work from Navarrete and Blat (2007) and Zhu, Hu, Janowicz, and McKenzie (2016). The instance-level conflation aims at merging the specific place entries in different gazetteers. There exist a variety of methods for measuring the similarities of spatial footprints (geometries) (Goodchild & Hunter, 1997; L. Li & Goodchild, 2011), place types (ontologies) (Rodríguez, Egenhofer, & Rugg, 1999), and place names (strings) (Sankoff & Kruskal, 1983). These similarity metrics are sometimes combined into workflows to conflate place entries, such as the work from Samal, Seth, and Cueto 1 (2004), Sehgal, Getoor, and Viechnicki (2006), and Hastings (2008).

There are other research topics in digital gazetteers. One is to equip digital gazetteers with the capability of reasoning. A digital gazetteer is typically implemented as a plain place dictionary without the ability to infer additional information from the existing content. Janowicz and Keßler (2008) and Keßler, Janowicz, et al. (2009) consider the place types in digital gazetteers as ontologies, and use logics to formally define and encode reasoning rules. They also designed prototypical Web and programming interfaces that can support subsumption (i.e., identifying the sub concepts of a broader concept) and similarity based reasoning. With rapid advances of Semantic Web technologies (see more in section 2.4), new data models and computational methods can enhance the query-answer capability of existing gazetteers. GeoNames is among the early pioneers that employed the Semantic Web technologies (Bizer, Heath, Idehen, & Berners-Lee, 2008). Another research topic in digital gazetteer focuses on the temporal dimension of places. These studies are often conducted in the context of historical gazetteers in which the place names, boundaries, or place types change over the years (Martins, Manguinhas, & Borbinha, 2008; Southall, Mostern, & Berman, 2011). A unique approach is presented by Mostern and Johnson (2008) who use events instead of locations as the fundamental unit for building a historical digital gazetteer.

## 2.3 Geographic Information Retrieval

Geographic information retrieval (GIR) is about retrieving relevant geographic information based on user queries (C. B. Jones & Purves, 2008). More generally, GIR can refer to retrieving geographic information from any type of data source, such as a structured database. However, studies on GIR usually focus on retrieving geographic information from unstructured data (Larson, 1996), especially from natural language text on the Web (McCurley, 2001; R. S. Purves, et al., 2007). It is estimated that 13% to 15% of Web queries contain place names (R. Jones, Zhang, Rey, Jhala, & Stipp, 2008; Sanderson & Kohler, 2004).

While GIR is traditionally considered an extension of information retrieval (Baeza-Yates & Ribeiro-Neto, 1999), it has also received a lot of attention from the GIScience community, as demonstrated by the GIR workshop series which began in 2004 (C. B. Jones & Purves, 2014) as well as the specialist meeting on Spatial Search held in Santa Barbara in 2015 (Ballatore, Hegarty, Kuhn, & Parsons, 2015).

It is not difficult to see the inherent connection between GIR and geospatial semantics. In order to retrieve relevant results, it is critical to *understand* the meaning of both the user queries and the candidate results (Janowicz, Raubal, & Kuhn, 2011). One important topic in GIR is place name disambiguation (also called toponym disambiguation) which aims at understanding the actual geographic place a place name refers to. Different places can have the same name (e.g., there are more than 43 populated places in the U.S. named "Washington"), and one place can have several different names (e.g., California is also called "The Golden State"). Therefore, how can we identify the correct geographic place when a place name shows up in a query or in a Web document? A general strategy towards place name disambiguation is to measure the similarity between the current context of the place name (i.e., the surrounding words) and the likely context of each possible candidate place. The likely context of the candidate places (e.g., the persons, organizations, or other places that are likely to be associated the place) can be extracted from external gazetteers or knowledge bases, such as WordNet (Davide Buscaldi & Paulo Rosso, 2008) and DBpedia (Y. Hu, Janowicz, & Prasad, 2014). These likely contexts can also be learned from data sources, such as Wikipedia, in a data-driven fashion (Cucerzan, 2007; Overell & Rüger, 2008). Similarity metrics, such as conceptual density (Agirre & Rigau, 1996) and cosine similarity (Bunescu & Pasca, 2006), can then be employed to quantify the similarities between the surrounding context of the place name and the likely context of each candidate place. While some similarity metrics are based on words and entities, some others are based on the geographic distance or overlap between the locations found in the surrounding context and the location of each candidate place (Davide Buscaldi & Paulo Rosso, 2008; Leidner, 2008; D. A. Smith & Crane, 2001). More recently, topic modeling techniques, such as Latent Dirichlet Allocation (LDA), have also been introduced to address the problem of place name disambiguation. For example, Ju, et al. (2016) proposed an approach that divides the surface of the Earth using a triangular mesh (see Figure 2).

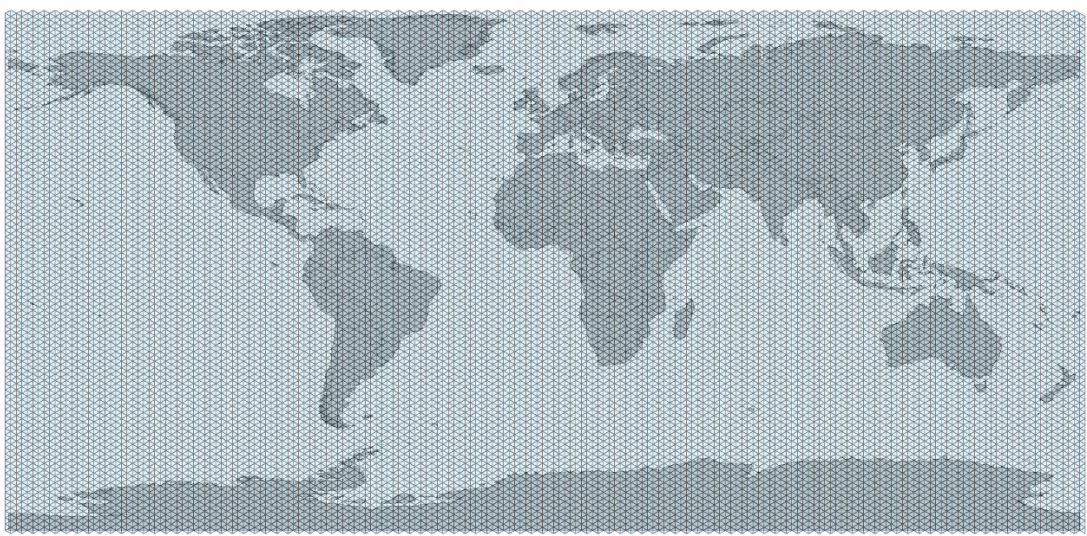

Figure 2. A triangular mesh for computing the thematic topics on the surface of the Earth.

Their approach then maps georeferenced Wikipedia texts into these triangles, and uses LDA to model the topics of each triangle. These topics are then compared with the target place name for disambiguation.

The problem of place name disambiguation is also closely related to *geoparsing* (Gelernter & Balaji, 2013; Gelernter & Mushegian, 2011; Moncla, Renteria-Agualimpia, Nogueras-Iso, & Gaio, 2014; Vasardani, Winter, & Richter, 2013; Wallgrün, et al., 2014), which involves first detecting the existence of place names from natural language text and then disambiguate it.

Another topic in GIR is ranking candidates based on the input query. Such a topic often boils down to computing the matching score between the input query and a candidate result. Once the matching scores are computed, the candidates can then be ranked. Many queries in GIR can be characterized using the format "<theme> <spatial relationship> <location>" (C. B. Jones & Purves, 2008), such as the query "natural disasters in California". Accordingly, the matching scores can be calculated based on these three components (Mata, 2007). To quantify the similarity between the thematic concepts in the query and those in the candidates, domain ontologies are often employed to identify the shortest distance between two concepts in an ontology (e.g., "natural disaster" and "earthquake") (C. B. Jones, Alani, & Tudhope, 2001). The thematic similarity can also be quantified through keyword expansion which enriches the input queries, which are typically short, with thematically related terms. External knowledge bases, such as WordNet (Buscaldi, Rosso, & Arnal, 2005; Y. Hu, Janowicz, Prasad, & Gao, 2015; Stokes, Li, Moffat, & Rong, 2008), or data mining approaches, such as latent semantic analysis (W. Li, Goodchild, & Raskin, 2014), can be employed for keyword expansion. Locations can be extracted from Web documents through geoparsing and place name disambiguation, and can then be used for spatially indexing the Web documents (Amitay, Har'El, Sivan, & Soffer, 2004; Silva, Martins, Chaves, Afonso, & Cardoso, 2006; C. Wang, Xie, Wang, Lu, & Ma, 2005). The extracted locations provide basis for computing the spatial relationships between the input query and the possible candidates. Spatial similarity metrics, such as those based on minimum bounding box (MBB) or convex hull (Frontiera, Larson, & Radke, 2008), have been used to quantify the spatial relations. In addition to the geometry-based approaches, digital gazetteers, which contain information about places and their relationships, are also used to quantify the similarity of places based on their proximity in the gazetteer graph (C. B. Jones, et al., 2002; Keßler, Janowicz, et al., 2009). With the similarity scores based on theme, spatial relationship, and location, a final matching score can then be computed by combining these scores. To evaluate the performance of different GIR methods, there are projects, such as GeoCLEF (Gey, et al., 2005) and the Toponym Resolution Markup Language (Leidner, 2006), which establish ground truth data by providing standard and annotated corpus. The frequently used evaluation metrics include precision, recall, and F1 score (Manning, Raghavan, & Schütze, 2008), which are defined as follows:

$$Precision = \frac{|Retrieved\ Relevant\ Results|}{|All\ Retrieved\ Results|} \quad (1)$$

$$Recall = \frac{|Retrieved\ Relevant\ Results|}{|All\ Relevant\ Results|} \quad (2)$$

$$F1\ score = 2 \cdot \frac{Precision \cdot Recall}{Precision + Recall} \quad (3)$$

In addition to retrieving geographic information from the Web, GIR is applied in many other contexts. One major application domain is spatial data infrastructure. As SDIs usually manage a large amount of data and metadata, it is important to use an effective search method that allows potential data users to quickly find the data they need (Janowicz, Wilkes, & Lutz, 2008). Ontologies are often used to associate metadata with external concepts and terminologies to facilitate data discovery. The application examples include Geosciences Network (GEON) (Bowers, Lin, & Ludascher, 2004), Linked Environments for Atmospheric Discovery (LEAD) (Droegemeier, et al., 2005), and Virtual Solar Terrestrial Observatory (VSTO) (Fox, et al., 2009). Another type of GIR applications focuses on modeling the task of the user. Instead of having a short keyword query, the user may have a more complicated task. Modeling the task of the user can help retrieve the information that can fit the user's needs. For example, Wiegand and García (2007) proposed to model professional tasks (e.g., emergency response) as ontologies, and identify the information that can be paired with the tasks. Y. Hu, Janowicz, and Chen (2016) modeled the spatiotemporal properties of the daily tasks of an individual (e.g., traveling to the workplace), and identified the geographic information that can help complete the task using information value theory. There are also GIR studies that focus on the design of effective query interfaces (C. B. Jones, et al., 2002; R. S. Purves, et al., 2007), as well as the searching and indexing of remote sensing images (Shyu, et al., 2007).

## 2.4 Geospatial Semantic Web and Linked Data

The Semantic Web was originally proposed by Berners-Lee, et al. (2001). It was a vision in which the Web was populated with structured and semantically-annotated data that can be consumed by not only human users but also machines. The Semantic Web can be considered as an enhancement of the existing document-based Web, on which the unstructured and natural language contents in Web pages are difficult for machines to understand. For example, a computer program may find it difficult to know that the current Web page is about *Washington, D.C.*, and that this page is linked to another page describing *the United States* is because *Washington, D.C.* is the capital of *the United States*. The vision of the Semantic Web was quickly embraced by the GIScience community (Egenhofer, 2002), and has influenced the thinking of the community on data organization, sharing, reusing, and answering complex spatial queries (Hart & Dolbear, 2013; Kuhn, Kauppinen, & Janowicz, 2014). The term *geospatial Semantic Web* has been frequently used in the studies that focus on the geospatial part of the Semantic Web (Y. Bishr, 2006; F. Fonseca, 2008; Yue, 2013).

Realizing the vision of the Semantic Web requires the current Web to be populated with structured and semantically-annotated data. Linked Data has been proposed by the World Wide Web Consortium (W3C) as a general guidance for publishing such data (Bizer, Heath, et al., 2009; Heath & Bizer, 2011). The term *Linked Data* has a two-fold meaning that is often used interchangeably. On the one hand, it refers to four principles for publishing well-structured data, such as using Uniform Resource Identifiers (URIs) and providing data descriptions readable to both humans and machines. On the other hand, it refers to the data that have been published following these four principles. Since the data to be published can come from a variety of domains with diverse attributes, Resource Description Framework (RDF) has been employed for organizing and publishing Linked Data. RDF represents data as *subject*, *predicate*, and *object*, and the three together is called a *triple* (Brickley & Guha, 2000; Hitzler, et al., 2009). Different data formats can be used to implement RDF, such as XML, Turtle, RDFa, and JSON-LD (Adams, 2016). From 2007 to 2014, more than 570 datasets (with billions of RDF triples) were published on the current Web (Y. Hu & Janowicz, 2016), forming the Linked Open Data (LOD) cloud which can be considered as a prototypical realization of the Semantic Web vision. Some of these Linked Data sets focus on geographic content, including GeoNames, Linked Geo Data (which contains OpenStreetMap data) (Auer, Lehmann, & Hellmann, 2009), and the Alexandria Digital Library Gazetteer (see Figure 3). Other Linked Data sets provide more general content, but also contain a significant amount of geographic data, such as DBpedia which is the Semantic Web version of Wikipedia (Auer, et al., 2007; Bizer, Lehmann, et al., 2009; Lehmann, et al., 2015). Some Linked Data sets are contributed by authoritative agencies, such as the U.S. Geological Survey (Usery & Varanka, 2012) and the U.K. Ordnance Survey (Goodwin, Dolbear, & Hart, 2008).

Two broad topics are investigated in geospatial Semantic Web and Linked Data: 1) how to effectively annotate and publish geospatial content? and 2) how to retrieve data to answer complex questions? For the first topic, geo-ontologies and ontology design patterns, such as the semantic sensor network ODP (Compton, et al., 2012) and the cartographic scale ODP (Carral, et al., 2013) discussed previously, have been developed to formalize the semantics of the data. Software tools, such as Triplify (Auer, Dietzold, Lehmann, Hellmann, & Aumueller, 2009), CSV2RDF (Ermilov, Auer, & Stadler, 2013), and TripleGeo (Patroumpas, Alexakis, Giannopoulos, & Athanasiou, 2014), have been developed to extract data from traditional data sources (e.g., relational databases) and convert them into RDF. Linked Data servers (also called triplestores), such as Virtuoso (Erling & Mikhailov, 2010) and GraphDB (Bellini, Nesi, & Pantaleo, 2015), can publish billions of triples and have provided some support to geospatial data. Despite these developments, questions, such as how to convert raster data into Linked Data and how to represent the temporal validity of geospatial data triples, still remain (Janowicz, Scheider, Pehle, & Hart, 2012; Kuhn, et al., 2014). For retrieving data from the Semantic Web, the SPARQL language (Pérez, Arenas, & Gutierrez, 2006) is a standard RDF query language recommended by W3C to retrieve Linked Data. The syntax of SPARQL is similar to Structure Query Language (SQL) that has been widely used in relational databases. However, SPARQL does not directly support the queries based on spatial relations (e.g., *within* and *intersects*). To address this issue, GeoSPARQL was proposed as an extension of SPARQL, and has been endorsed by the Open Geospatial Consortium (OGC) (Battle & Kolas, 2011). Triplestores, such as Parliament (Battle & Kolas, 2012) and Oracle Spatial and Graph (Perry, Estrada, Das, & Banerjee, 2015), have also implemented GeoSPARQL to support geospatial queries. In addition to GeoSPARQL, there exist other research efforts for supporting spatial and temporal queries, such as the work from Perry, Sheth, Hakimpour, and Jain (2007) and Gutierrez, Hurtado, and Vaisman (2005).

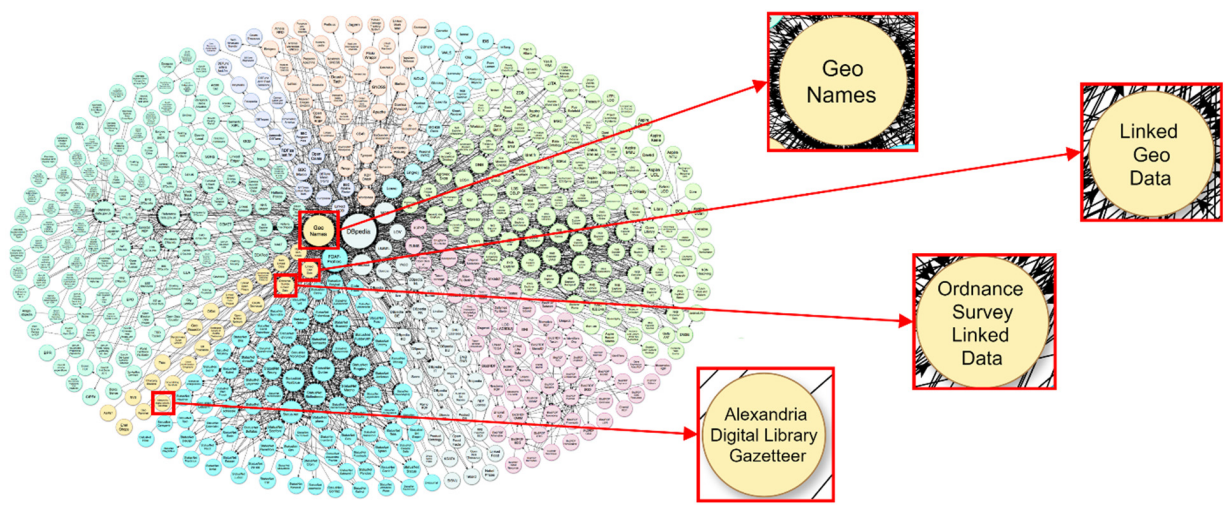

Figure 3. The Linked Open Data cloud as of August 30, 2014 (http://lod-cloud.net/), and several major geographic data sets.

Many other applications make use of the technologies and data models from the geospatial Semantic Web. One type of applications is spatial data infrastructures which can be considered as local geospatial Semantic Webs. The Linked Data principles have been applied to interlinking the metadata and services hosted by a SDI, and have facilitated search and resource discoveries (Athanasis, Kalabokidis, Vaitis, & Soulakellis, 2009; Janowicz, et al., 2010; Zhang, Zhao, Li, & Osleeb, 2010; Zhao, Zhang, Anselin, Li, & Chen, 2015). Another type of applications is using Linked Datasets as external knowledge bases to support named entity recognition and disambiguation. The examples include DBpedia Sportlight (Mendes, Jakob, García-Silva, & Bizer, 2011) and Open Calais (Gangemi, 2013) which can identify and extract geographic places and other types of entities (e.g., persons, companies, and universities) from unstructured texts with high accuracy and computational efficiency. There are also Linked Data driven portals that enable users to interactively explore Linked (Geo) Data by following the links between entities. One example is *Spatial@LinkedScience* which hosts the bibliographic Linked Data for researchers, papers, and organizations in major GIScience conferences (Keßler, Janowicz, & Kauppinen, 2012). Another example is the portal developed by Mai, Janowicz, Hu, and McKenzie (2016) which enables users to explore Geoscience and Oceanography Linked Data from tabular, graph, and map views.

## 2.5 Place Semantics

*Place* is an important concept in GIScience (Goodchild, 2011; Winter, Kuhn, & Krüger, 2009). The notion of *place* is closely associated with human experience, and can be differentiated from *space* (Couclelis, 1992; Fisher & Unwin, 2005; Tuan, 1977). Accordingly, not any space can be considered as a place. Place plays an indispensable role in human communication, and place names are frequently used in daily conversations (Cresswell, 2014; Winter & Freksa, 2012). Research on place semantics focuses on understanding the meaning of places through human descriptions and human-place interactions. Traditionally, interviews were employed to elicit people's opinions toward places (Cresswell, 1996; Montello, Goodchild, et al., 2003). While such an approach sheds valuable insight on human experience, they were labor-intensive and therefore may not be suitable for the studies that involve many places over a multi-year timespan. The emergence of the Web, especially Web 2.0, brings a big volume of data, and a lot of these data are contributed by the general users (Goodchild, 2007; Stoeckl, Rohrmeier, & Hess, 2007). Place descriptions and human-place interactions are often contained in Web data which offer great opportunities for studying places (R. Purves, Edwardes, & Wood, 2011; Winter, et al., 2009). By harvesting place-related data and designing automatic algorithms, we can perform studies that can scale up. This section focuses on data-driven and computational approaches towards place semantics.

Two types of place-related data can be found on the Web. The first type contains only textual descriptions about places. Examples include place descriptions on city websites (Tomai & Kavouras, 2004b), travel blogs, and Wikipedia articles. This type of data does not contain explicit geographic coordinates, and geoparsing is necessary to extract and locate places (Vasardani, et al., 2013). Wikipedia is a valuable data resource that provides descriptions about a large number of cities and towns in the world. The second type of data contains associations between descriptions and geographic coordinates. Examples include the various location-based social media data, such as geotagged tweets and Flickr photos. With the given link between descriptions and locations, the descriptive texts can be aggregated to the corresponding locations for place studies. Usually, location-based social media data also contain the time when a user interacts with a place (e.g., by checking in), and therefore can be used to study human-place interactions. In addition to the wide availability of place-related data, the fast development of natural language processing (NLP) techniques and standard tools have significantly boosted the research in place semantics. For example, Manning, et al. (2014) developed the Stanford NLP Toolkit which requires little programming background of the users. The text mining package in R (the *tm* package) also lowers the barrier of performing text analysis, especially for researchers who are already familiar with R (Feinerer & Hornik, 2012; Meyer, Hornik, & Feinerer, 2008).

Place semantics can be studied from thematic, spatial, and temporal perspectives. The thematic perspective examines human experiences toward places through natural language descriptions. In many cases, the thematic topics discussed by people about or at a place are related to the activities that can be afforded by the place, reflecting Gibson's affordance theory (Gibson, 1982). These thematic topics can be revealed through simple approaches, such as word clouds. Figure 4(a) and (b) show the top 20 most frequent words from the reviews of two places (stop words, such as "the", "of", and "in", are removed). Without any further information, one can easily tell the general place type and the supported activities. More advanced NLP methods, such as Latent Dirichlet Allocation (Blei, Ng, & Jordan, 2003), are also used in recent years to model thematic topics based on place-related data. For example, Adams, McKenzie, and Gahegan (2015) and Adams (2015) have proposed LDA-based approaches for extracting thematic topics for places, quantifying place similarity thematically, and searching places based on topics (e.g., finding places associated with the topic *beach*). Emotions, as part of human experience, can also be extracted through sentiment analysis to understand places thematically (Ballatore & Adams, 2015). Space is another perspective for studying places. This perspective often focuses on representing the vague boundary of places-- a topic that has been discussed in the previous section Digital Gazetteers. There is also research on spatially representing places based on the surrounding landmarks, and examples include the works from Kim, Vasardani, and Winter (2016) and Zhou, Winter, Vasardani, and Zhou (2016). These studies do not attempt to construct geometries (e.g., polygons) or density surfaces to represent the spatial footprints of places; instead, they formalize the spatial relations between the target place and nearby landmarks into a graph representation. The third perspective for studying place semantics takes a temporal view. This perspective examines the time when people are more (or less) likely to interact with a place (Ye, Shou, Lee, Yin, & Janowicz, 2011). For example, McKenzie, Janowicz, Gao, Yang, and Hu (2015) examined the check-in data of Foursquare users at different types of Points Of Interest (POI), and used the temporal check-in patterns to characterize places. Such temporal patterns also reflect human activities at the corresponding places (e.g., *restaurants* typically receive high number of check-ins during lunch and dinner hours), and can be applied to other applications, such as reverse geocoding (McKenzie & Janowicz, 2015).

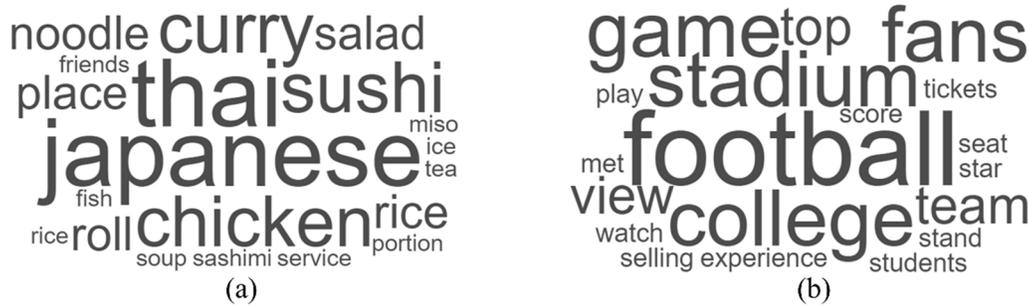

Figure 4. Word clouds constructed based on the reviews of two places.

In addition to the three unit perspectives, place semantics can be studied by combining two or all three of them. For example, by combining space and time, we can examine the spatiotemporal dynamics of places, such as how the vague boundaries of places change over the years. By combining space and theme, we can develop topic maps that visualize the major topics related to different locations (Kennedy, Naaman, Ahern, Nair, & Rattenbury, 2007; Rattenbury & Naaman, 2009). By combining time and theme, we can explore the evolution of topics over time, such as the emergence of new topics and the disappearance of old ones. There are also studies that combine space, time, and thematic topics to obtain a more comprehensive understanding on places and the associated events (Y. Hu, Gao, et al., 2015; W. Wang & Stewart, 2015). With the availability of new place-related data sources, such as Yik Yak (McKenzie, Adams, & Janowicz, 2015), machine learning methods, such as deep learning (LeCun, Bengio, & Hinton, 2015), and data mining tools, such as TensorFlow (Abadi, et al., 2015), more research on place semantics is waiting to be conducted.

**2.6. Cognitive Geographic Concepts and Qualitative Reasoning**

Cognitive geographic concepts generally refer to the informal geographic knowledge that people acquire and accumulate during the interactions with the surrounding environment (Golledge & Spector, 1978). Such informal knowledge was termed as *naïve geography* by Egenhofer and Mark (1995), and can be differentiated from the formal geographic knowledge that requires intentional learning and systematical training (Golledge, 2002). The training requirements of formal geographic knowledge can be seen from the special concepts and terminologies, such as projected coordinate systems, raster, vector, and map algebra, discussed in many GIS textbooks (Bolstad, 2005; K. C. Clarke, 1997; Longley, Goodchild, Maguire, & Rhind, 2001). Since not every GIS user has received formal training, understanding the conceptualization of general people towards geographic concepts can facilitate the design of geographic information systems (B. Smith & Mark, 1998). This section will focus on the informal understanding of general people towards geographic concepts and spatial relations, given the focus of this chapter on geospatial semantics. However, it is worth noting that the content discussed in this section is only part of the field of cognitive geography (Montello, 2009; Montello, Fabrikant, Ruocco, & Middleton, 2003) which involves many other topics such as geo-visualization (MacEachren & Kraak, 2001).

Geographic concepts and spatial relations are two types of informal geographic knowledge that people develop in everyday life. Studies on the former often examine the *typical examples* that people associate with the corresponding geographic concepts. For example, B. Smith and Mark (2001) found that non-expert individuals usually think about entities in the physical environment (e.g., *mountains* and *rivers*) when asked to give examples of *geographic features or objects*; whereas they are more likely to answer with social or built features (e.g., *roads* and *cities*) when asked *things that could be portrayed on a map*. Such *typical examples* can be explained by the prototype theory from E. H. Rosch (1973) and E. Rosch and Lloyd (1978) in psychology, in which some members are better examples of a category. For example, *robin* is generally considered as a better example for the category of *bird* compared with *penguin*. There are values in understanding the typical examples of geographic concepts. For instance, it can help increase the precision of geographic information retrieval by identifying the default geographic entities that are more likely to match the search terms of a user. Besides, it has been found that different communities, especially the communities with different languages, may establish their own conceptual systems (D. M. Mark & Turk, 2003). Understanding these conceptualization differences on geographic concepts can help develop GIS that can better fit local needs (B. Smith & Mark, 2001).

Spatial relation is another type of informal geographic knowledge that we acquire by interacting with the environment. Figure 5 provides an example which illustrates the spatial relations that a person may develop for different places near the campus of the University of California Santa Barbara (UCSB). Such spatial relations are qualitative: we may know the general locations and directions of these places but not the exact distances between them (e.g., the distance between *Costco* and *Camino Cinemas* in meters is unknown to the person). Yet, these informal spatial relations are useful and sufficient for many of our daily tasks such as wayfinding and route descriptions (Brosset, Claramunt, & Saux, 2007; Klippel, Tappe, Kulik, & Lee, 2005; Klippel & Winter, 2005; Montello, 1998). In addition, these relations are convenient to acquire since we do not always carry a ruler to measure the exact distances and angles between objects. These informal spatial relations also present an abstraction from some quantitative details and are not restricted to a set of specific values (e.g., the spatial relation *A is to the west of B* can represent an infinite number of *A* and *B*, as long as they satisfy this relative spatial constraint) (Freksa, 1991; Gelsey & McDermott, 1990).

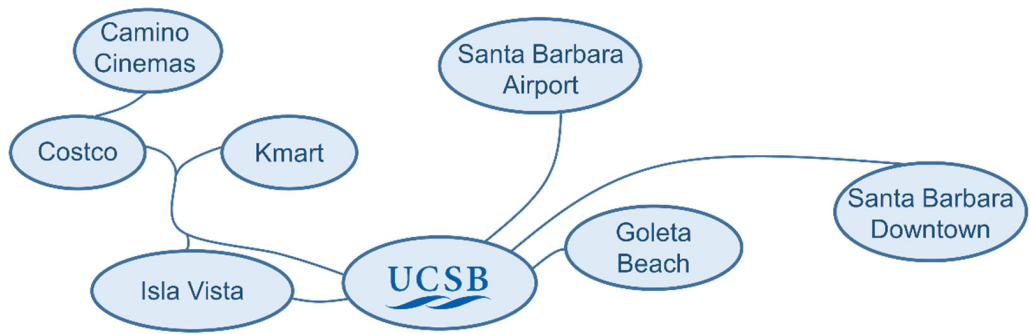

Figure 5. Qualitative relations a person may develop for the places around UCSB.

The informal conceptualization of people on geographic concepts and spatial relations can be formally and computationally modeled to support qualitative reasoning. The term *qualitative reasoning* should be differentiated from the term *qualitativeness* which may imply descriptive rather than analytical methods (Egenhofer & Mark, 1995). Spatial calculi can be employed to encode spatial relations (Renz & Nebel, 2007), such as the mereotopology (B. L. Clarke, 1981), 9-intersection relations (Egenhofer, 1991; Egenhofer & Franzosa, 1991), double-cross calculus (Freksa, 1992), region connection calculus (Randell, Cui, & Cohn, 1992), flip-flop calculus (Ligozat, 1993), and cardinal direction calculus (Frank, 1996). In addition to spatial relations, temporal relations, such as the relative relations between events, can also be formally represented using, e.g., the interval algebra proposed by Allen (1983). Algorithmically, informal knowledge can be modeled as a graph with nodes representing geographic concepts (and their typical instances) and edges representing their spatial relations. If we restrict the nodes to be only place instances, we can derive a *place graph*. This graph representation is fundamentally different from the Cartesian coordinates and geometric rules widely adopted in existing geographic information systems, and could become the foundation of *place-based GIS* (Goodchild, 2011). Platial operations, as counterparts of the spatial operations (Gao, Janowicz, MckKnzie, & Li, 2013), could then be developed by reusing and extending the existing graph-based algorithms as well as designing new ones. Constructing such a geographic knowledge graph can be challenging, since different individuals often conceptualize places and spatial relations differently. However, such a challenge also brings the opportunity of designing more personalized GIS for supporting the tasks of individuals (Abdalla & Frank, 2011; Abdalla, Weiser, & Frank, 2013). Qualitative reasoning and place-based GIS should not be seen as replacements for quantitative reasoning and geometry-based GIS (Egenhofer & Mark, 1995). Instead, they complement existing methods and systems, and should be used when the application context is appropriate.

## 3. Summary and Outlook

This chapter has provided a synthetic review on the research related to geospatial semantics. Specifically, six major areas are identified and discussed, including semantic interoperability, digital gazetteers, geographic information retrieval, geospatial Semantic Web, place semantics, and cognitive geographic concepts. For content organization, each research area is discussed as a separate section. However, they are interconnected and can be involved simultaneously in one study. For example, research on geospatial Semantic Web usually also employs ontologies to facilitate semantic interoperability. Similarly, digital gazetteers are widely used in GIR and place semantics to extract and disambiguate place names. In addition, understanding naïve geographic concepts is important for developing ontologies that can be agreed by multiple communities.

The six major research areas share a common core, namely understanding the meaning of geographic information. Such an understanding brings its own values to geospatial research and applications. For example, constructing ontologies, gazetteers, and Linked Data can help machines process geographic information more effectively and automatically extract knowledge more efficiently. Examining place semantics can help people quickly grasp the meaning of places from a large amount of texts. Modeling cognitive geographic concepts and designing intelligent GIR algorithms can help machines understand the thinking of people, thus facilitating the interactions between GIS and users. In sum, geospatial semantics offers a unique semantic perspective towards advancing GIScience research.

Other topics in geospatial semantics can be investigated in the near future. For example, the studies reviewed in this chapter focused on English data, and there is a lot of geographic information collected and recorded in other languages. Research efforts, such as those from Ouksel and Sheth (1999), D. M. Mark and Turk (2003), Nowak, Nogueras-Iso, and Peedell (2005), and Mata-Rivera, Torres-Ruiz, Guzmán, Moreno-Ibarra, and Quintero (2015), are important for understanding and processing multilingual and multicultural datasets. In addition, there can be more applications of high-performance computing (HPC) in geospatial semantics. Existing HPC frameworks (S. Wang, 2010; Yang, et al., 2011) can be integrated with the semantic methods to crunch the large volume of geospatial and textual data. Given the rich amount of existing literature, this chapter cannot cover all related studies. However, hopefully it can serve as an entry point for exploring the world of geospatial semantics.